\documentclass[preprint]{article}

    \PassOptionsToPackage{numbers, compress}{natbib}

\usepackage{float}

\usepackage[preprint]{neurips_2024}

\usepackage[utf8]{inputenc} 
\usepackage[T1]{fontenc}    
\usepackage{hyperref}       
\hypersetup{
    colorlinks=false,
    pdfborder={0 0 0},
    pdfborderstyle={/S/U/W 0},
}
\usepackage{url}            
\usepackage{booktabs}       
\usepackage{amsfonts}       
\usepackage{nicefrac}       
\usepackage{microtype}      
\usepackage{xcolor}         
\usepackage{graphicx}
\usepackage{listings}
\lstdefinestyle{mystyle}{
    backgroundcolor=\color{gray!10},
    commentstyle=\color{gray},
    keywordstyle=\normalfont,
    basicstyle=\ttfamily\footnotesize,
    breaklines=true,
    captionpos=b,
    numbersep=5pt,
    showspaces=false,
    showstringspaces=false,
    showtabs=false,
    tabsize=2
}
\usepackage{amssymb}
\usepackage{pifont}
\usepackage{courier}
\usepackage{tcolorbox}
\tcbuselibrary{listingsutf8}
\newtcblisting{codeblock}{
  listing only,
  colback=gray!10,
  colframe=white,
  listing options={
    basicstyle=\ttfamily\footnotesize,
    keywordstyle=\normalfont,
    breaklines=true,
    captionpos=b,
    extendedchars=true,
    showtabs=false,
    showspaces=false,
    showstringspaces=false,
    tabsize=2,
    inputencoding=utf8,
    literate={✗}{{\ding{55}}}1
             {×}{{\ensuremath{\times}}}1,
  }
}

\title{TaxCalcBench: Evaluating Frontier Models on the Tax Calculation Task}

\author{Michael R. Bock\And Kara Molisee\And Zachary Ozer\And Sumit Shah \AND
Column Tax}

\begin{document}

\maketitle

\begingroup
  \renewcommand\thefootnote{}       
  \footnotemark\footnotetext{Code and data: \url{https://github.com/column-tax/tax-calc-bench}.}
  \addtocounter{footnote}{-1}       
\endgroup

\begin{abstract}
  Can AI file your taxes? Not yet. Calculating US personal income taxes is a task that requires building an understanding of vast amounts of English text and using that knowledge to carefully compute results. We propose TaxCalcBench, a benchmark for determining models' abilities to calculate personal income tax returns given all of the necessary information. Our experiment shows that state-of-the-art models succeed in calculating less than a third of federal income tax returns even on this simplified sample set. Our analysis concludes that models consistently misuse tax tables, make errors in tax calculation, and incorrectly determine eligibility. Our findings point to the need for additional infrastructure to apply LLMs to the personal income tax calculation task.
\end{abstract}

\begin{figure}[h!]
  \centering
  {\includegraphics[width=.8\linewidth]{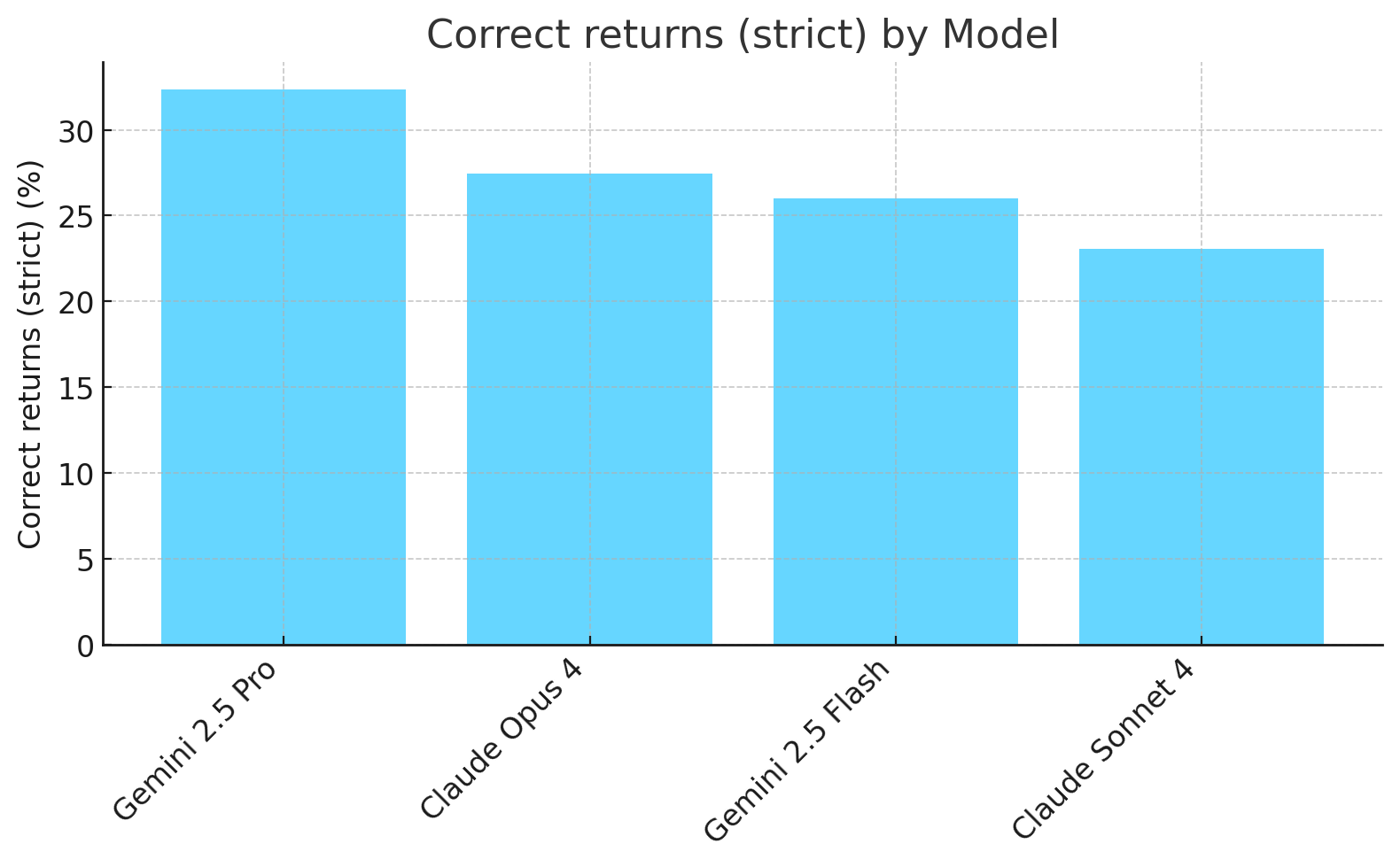}}
\end{figure}

\section{Introduction}

LLMs have become increasingly capable over the past year at coding and math tasks thanks to improvements in reasoning via reinforcement learning. This increase in coding, math, and reasoning ability is evidenced by frontier models' performance improvements on benchmarks like AIME, LiveCodeBench, Aider Polyglot, SWE-bench Verified, and TAU-bench. And while tax calculation has been used in fun LLM demos \cite{1}, we haven't seen LLMs formally tested on their ability to calculate taxes.

In this paper, we describe tax calculation and filing tasks, the TaxCalcBench benchmark we've created to test models on this task, how the benchmark was created, and the results of testing the latest frontier models on this task.

Tax filing is an exercise in once-a-year personal financial data collection. Once you collect all of your information, you (or your accountant) prepare and enter that information into tax filing software. Behind the scenes of that tax filing software is a ``tax engine'' that computes the income, tax liability, credits, etc. that make up your tax return.

TaxCalcBench aims to evaluate model performance on that third and final task: tax calculation. TaxCalcBench is a series of 51 test cases that represent a modest range of personal income tax returns. The test cases include the complete set of user inputs required to compute a tax return and the correct expected output from a traditional tax engine. The Tax Year 2024 (TY24) version of TaxCalcBench includes a set of federal-only tax returns representing just a share of Americans' tax\break situations.

Our experiment shows that frontier models cannot reliably calculate taxes. Even the best-performing model can only compute less than a third of returns correctly. When using a less precise evaluation criteria that allows for plus-or-minus \$5 of tax owed or refund due (which is not allowed in tax calculation, but interesting nonetheless), models get 15-20\% more returns correct on an overall\break basis.

Our analysis finds that models consistently use incorrect tax tables, make calculation errors, and incorrectly determine eligibility, leading to overall incorrectly computed tax returns.

Our findings point to a continued need for deterministic tax calculation engines to ensure accuracy and the need for additional infrastructure and orchestration to augment LLMs to be able to reliably compute tax returns.

\section{Background: The Tax Calculation Task}

Tax filing consists of 3 main subtasks:

\begin{enumerate}
\item  \textbf{Document collection}: collecting all of the documents (e.g. W-2s) required for filing.

\item  \textbf{Preparation}: entering all of the collected information into tax preparation software.

\item  \textbf{Calculation}: transforming the entered information into the completed tax return (\href{https://www.irs.gov/forms-pubs/about-form-1040}{Form 1040}, for personal income tax) for filing.
\end{enumerate}

This benchmark is solely focused on \textbf{(3) Calculation}.

To date, companies have built ``tax calculation engines'' as deterministic software: code that can compute the tax return given a user's information. Only about a dozen tax engines have ever been built, and very few in the past two decades \cite{2}.

Tax filing is an exercise in once-a-year personal financial data collection: taxpayers collect all of their financial data and facts for the year (e.g. W-2s, 1099s, dependents’ DoBs, spouse’s data if they're married) so that it can be entered into tax software either by the taxpayer themselves or by an accountant. Tax software developers call this data ``inputs''.

A tax engine takes these ``inputs'' and runs them through calculations and transformations defined by the IRS, state, and local agencies. The tax engine then outputs the completed tax return in the XML format specified by the IRS to be e-filed and a PDF for users to view their return.

The calculations and transformations from the IRS, state, and local agencies are defined only in English. The core challenge of building a traditional tax engine is translating those calculations from English into programmatic code that can compute a user’s tax return, 100\% accurately for every possible permutation of user inputs.\eject

One example is Line 1a of Form 1040: ``Total amount from Form(s) W-2, box 1 (see instructions)''. If the user has two W-2s, one with \$30k in box 1 and the other with \$20k in box 1, Form 1040 Line 1a will be the sum, \$50k:

\begin{figure}[h!]
  \centering\vspace*{-12pt}
  {\includegraphics[width=0.99\linewidth]{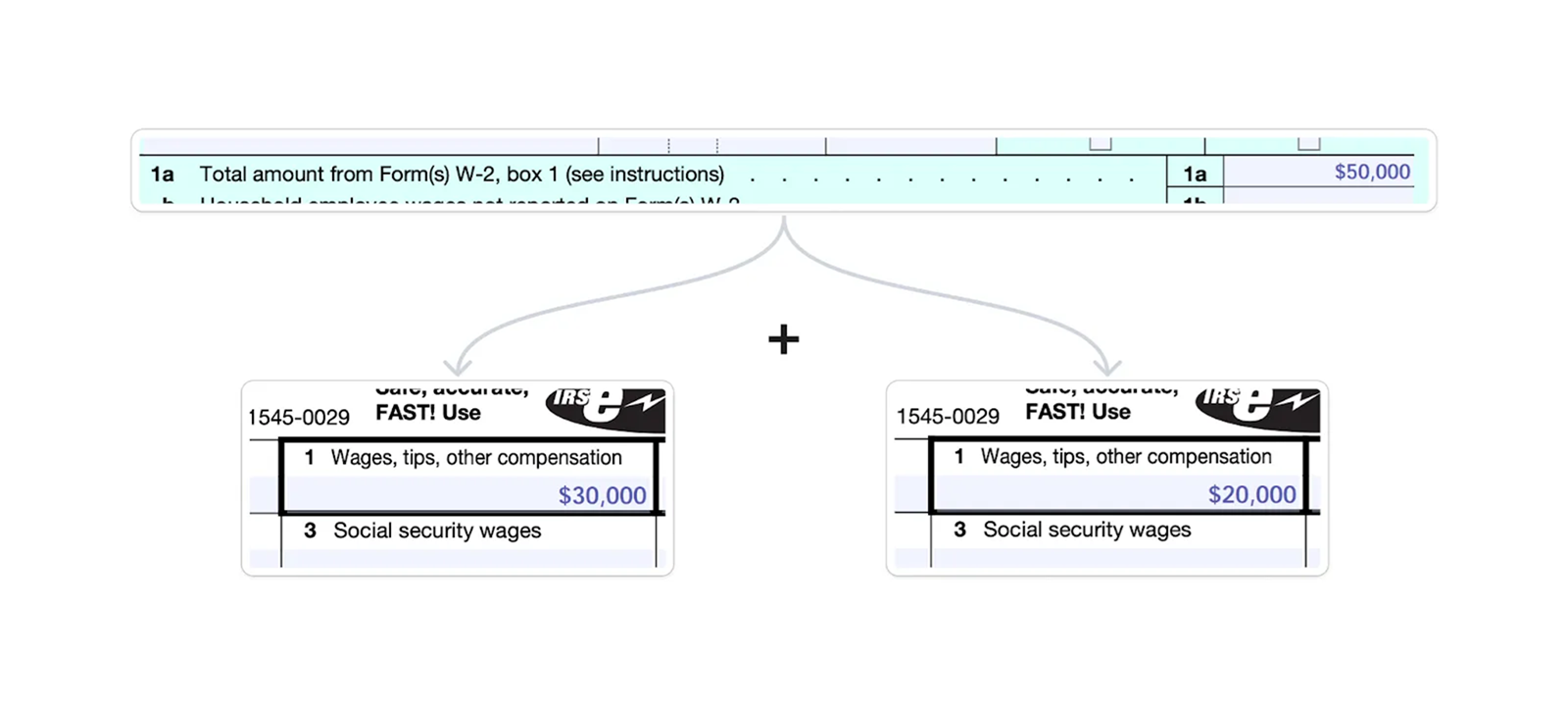}}\vspace*{-16pt}
\end{figure}

The code behind the scenes of that calculation might look something like:

\begin{lstlisting}[language=Python]
line_1a = sum(w2.box_1 for w2 in w2s) -
sum(sch_c.temporary_statutory_employee for sch_c
in schedules_c) - schedule_1.nonqualified_deferred_compensation
\end{lstlisting}

The ``(see instructions)'' parenthetical above is crucial in making this formula more complex than simply summing all box 1s across W-2s, which leads to the added references to \texttt{temporary\_statutory\_employee} and \texttt{nonqualified\_deferred\_compensation}.

Contrary to popular belief, the IRS doesn't ``know the answer'' ahead of time: they do not maintain their own first-party tax engine. As a result, this is something that each tax software provider must implement.

For a sense of scale, the US income tax code, across Federal and ${\sim}$40 states, is more than 75k pages and over a million lines of text total. It's not the arithmetic operations (mostly addition, subtraction, multiplication, division) themselves that are particularly difficult. Instead, it's the number of rules and the number of interconnections between those rules -- in addition to the fact that there is no given ``answer key'' from the IRS or States -- that makes this problem hard. A single input like Filing Status is a dependency with downstream impacts on thousands of additional calculations on the Form 1040 alone.

Traditional tax engines \cite{3} have built this computation graph by-hand. In this simplified diagram, each node like ``calculate'' and ``sum'' represents a single calculation like the Line 1a example above. These calculations are very interconnected and eventually produce the expected output (in XML and PDF formats):

\vspace{4pt}
\begin{figure}[hb!]
  \centering
    \vspace*{-24pt}
  {\includegraphics[width=.6\linewidth]{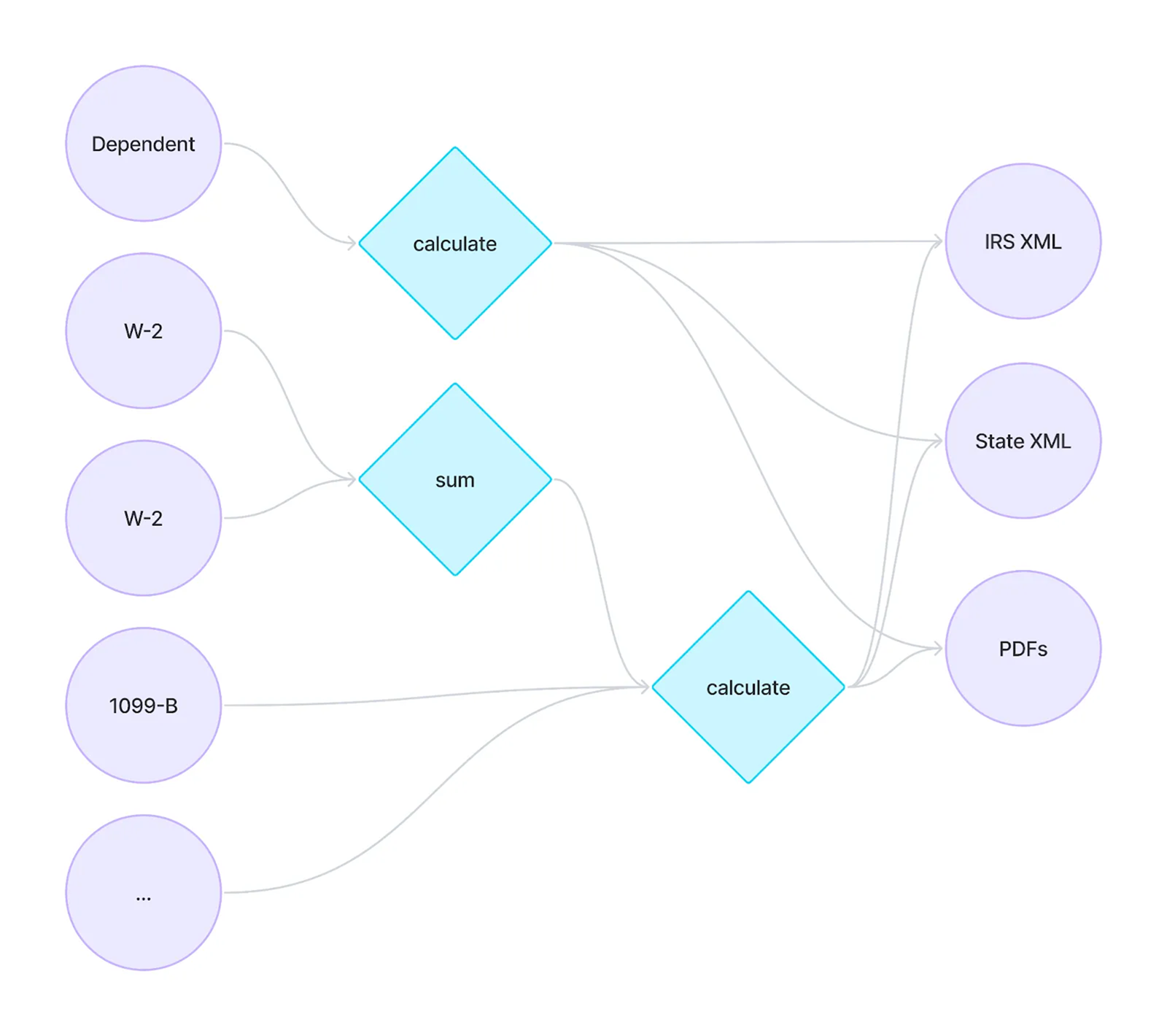}}
  \vspace*{-24pt}
\end{figure}\vfill\eject

\section{TaxCalcBench}

The TaxCalcBench eval is a dataset of 51 pairs of user inputs and the expected correctly-computed tax return output as well as a testing harness that compares models' output to the expected result.

\subsection{The TaxCalcBench dataset}

The dataset represents a mix of tax situations (income types, filing statuses, credits and deductions) for a fairly simple set of Federal-only tax returns (e.g. for users who live in non-income tax states like Florida and Texas).

This dataset is hard to come by: it's been crafted by hand by a team of human Tax Software Analyst  experts.

The inputs are formatted in a proprietary JSON. The inputs represent all of the information needed to fully calculate the output return. In other words, the \textbf{Document collection} and \textbf{Preparation} tasks (see Background section above) can be assumed to have been completed 100\% correctly.

A portion of the input representing a user's W-2s (shortened for clarity) looks like:

\begin{lstlisting}
"w2": [
  {
    "employer_name": {
      "label": "Employer's name",
      "value": "Acme Corp"
    },
    "wages": {
      "label": "Box 1",
      "value": 50000
    },
    "withholding": {
      "label": "Box 2",
      "value": 2000
    },
    "social_security_wages": {
      "label": "Box 3",
      "value": 50000
    },
    "social_security_tax": {
      "label": "Box 4",
      "value": 3100
    },
    "medicare_wages_and_tips": {
      "label": "Box 5",
      "value": 50000
    },
    "medicare_tax_withheld": {
      "label": "Box 6",
      "value": 725
    }
  }
]
\end{lstlisting}

The outputs are formatted as IRS-expected ``\href{https://www.irs.gov/e-file-providers/modernized-e-file-mef-schemas-and-business-rules}{Modernized e-File (MeF)}'' XML.

A portion of the output (shortened for clarity) looks like:

\begin{lstlisting}[language=XML]
<IRS1040 documentId="1">
  <IndividualReturnFilingStatusCd>1</IndividualReturnFilingStatusCd>
  <VirtualCurAcquiredDurTYInd>false</VirtualCurAcquiredDurTYInd>
  <TotalExemptPrimaryAndSpouseCnt>1</TotalExemptPrimaryAndSpouseCnt>
  <TotalExemptionsCnt>1</TotalExemptionsCnt>
  <WagesAmt referenceDocumentId="IRSW2-0">50000</WagesAmt>
  <WagesSalariesAndTipsAmt>50000</WagesSalariesAndTipsAmt>
  <TotalIncomeAmt>50000</TotalIncomeAmt>
  <AdjustedGrossIncomeAmt>50000</AdjustedGrossIncomeAmt>
  <TotalItemizedOrStandardDedAmt>14600</TotalItemizedOrStandardDedAmt>
  <TotalDeductionsAmt>14600</TotalDeductionsAmt>
</IRS1040>
\end{lstlisting}

This dataset consists of only Tax Year 2024 (TY24) returns. The dataset contains federal-only returns for fairly simple tax situations and includes features like:

\begin{itemize}
\item  Filing statuses: Single, Married Filing Jointly, and Head of Household

\item  Income sources: W-2, self-employed, capital gains, interest, and dividends

\item  Credits and deductions: Child Tax Credit, Earned Income Tax Credit, Child and Dependent Care Expenses
\end{itemize}

The input/output pairs are all named with descriptive tiles, e.g. \texttt{hoh-multiple-w2-box12-}\break\texttt{codes}, stored in \href{https://github.com/column-tax/tax-calc-bench/tree/main/tax_calc_bench/ty24/test_data}{a directory} including an \texttt{input.json} and \texttt{output.xml} file for each test.

\subsection{Benchmark Construction}

We constructed TaxCalcBench by utilizing a subset of the tests we use at \href{https://www.columntax.com/}{Column Tax} to test our tax calculation engine. These tests were created by human experts and are verified by a deterministic tax engine.

While these test cases represent tax return situations of many Americans, the actual data in these tests is totally synthetic: no real taxpayer data is represented in this test data at all.

\section{Experiment}

\subsection{Methodology}

TaxCalcBench tests models with a knowledge cutoff in 2025 on their ability to natively calculate a correct tax return for the 2024 Tax Year: Gemini 2.5 Pro, Gemini 2.5 Flash, Claude Opus 4, and Claude Sonnet 4.

The baseline for performance is 100\% because the models are being compared to a deterministic ``traditional code'' engine.

TaxCalcBench conducts its tests by prompting the model to calculate a tax return given the full set of user inputs. \href{https://github.com/column-tax/tax-calc-bench/blob/main/tax_calc_bench/tax_return_generation_prompt.py}{Here is the prompt} used, which asks the model to output the return in a simplified text-only format (\textit{not} the proper XML because models can't yet natively produce MeF schema-compatible XML\footnote{At the moment, the XML schemas are much too large to fit in context, and breaking them up to only include the relevant schemas would require an understanding of the tax calculations themselves and be quite a lot of scaffolding/infrastructure to provide the models.}):

\begin{lstlisting}
Form [NUMBER]: [NAME]
==================
Line 1: [Description] | [Explanation of calculations, if any] | [Amount]
Line 2: [Description] | [Explanation of calculations, if any] | [Amount]
...
\end{lstlisting}

\href{https://github.com/column-tax/tax-calc-bench/blob/main/tax_calc_bench/tax_return_evaluator.py}{The evaluator} then compares the [\texttt{Amount}]s generated by the model to the expected values in the output XML on a line-by-line basis for the most-important lines (the lines that most impact Tax owed and balance due/refund) of the main Form 1040 tax return.

For example, the model might output:

\begin{lstlisting}
Line 1a: Total amount from Form(s) W-2, box 1 | $32,456 + $15,444 | 47900
\end{lstlisting}

Which is then compared to the content of the proper XML tag (at XPath \texttt{/Return/}\break\texttt{ReturnData/IRS1040/WagesAmt}):

\begin{lstlisting}
<WagesAmt referenceDocumentId="IRSW2-0 IRSW2-1">47900</WagesAmt>
\end{lstlisting}

Each run is evaluated by:

\begin{itemize}
\item  Correct returns (strict): Model outputted returns are considered correct if the amounts strictly match for every evaluated line. This is the only actual metric that matters in the end because the IRS expects 100\%-correctly computed tax returns\footnote{There is occasionally some ambiguity with regard to rounding rules that could lead to multiple acceptable strict values.}.

\item  TaxCalcBench also evaluates and reports on these additional metrics that give additional color to the models' performances:

\begin{itemize}
\item  Correct returns (lenient): if every evaluated line is within +/- \$5 of the expected value.

\item  Correct (by line): the percent of evaluated lines that match the expected value.

\item  Correct (by line, lenient): the percent of evaluated lines that are within +/- \$5 of the expected value.
\end{itemize}
\end{itemize}

Models are evaluated at 5 thinking levels to determine if additional thinking budget is beneficial to their performance on the TaxCalculation task:

\begin{itemize}
\item  \texttt{lobotomized}: either no thinking token budget or the lowest thinking budget allowed by the model

\item  \texttt{low}: translates to \href{https://docs.litellm.ai/docs/providers/gemini#usage---thinking--reasoning_content}{OpenAI's reasoning\_effort} at 1024 budget thinking tokens

\item  \texttt{medium}: translates to \href{https://docs.litellm.ai/docs/providers/gemini#usage---thinking--reasoning_content}{OpenAI's reasoning\_effort} at 2048 budget thinking tokens

\item  \texttt{high}: translates to \href{https://docs.litellm.ai/docs/providers/gemini#usage---thinking--reasoning_content}{OpenAI's reasoning\_effort} at 4096 budget thinking tokens

\item  \texttt{ultrathink}: the highest thinking token budget allowed by the model
\end{itemize}

Additionally, TaxCalcBench includes 4 runs per model at each thinking level, allowing us to calculate pass@k and pass\string^k\footnote{pass\string^k is a metric of reliability over several runs defined by the $\tau$-bench paper\cite{7}.} metrics.

\subsection{Takeaways}

\begin{center}
\renewcommand{\arraystretch}{1.2}
\begin{tabular}{|l|c|c|c|c|} \hline
\textbf{Model} & \parbox{2cm}{\centering\textbf{Correct returns (strict)}} & \parbox{2cm}{\centering\textbf{Correct returns (lenient)}} & \parbox{2cm}{\centering\textbf{Correct (by line)}} & \parbox{2cm}{\centering\textbf{Correct (by line, lenient)}} \\ \hline
Gemini 2.5 Pro & 32.35\% & 51.96\% & 81.22\% & 86.12\% \\ \hline
Claude Opus 4 & 27.45\% & 42.65\% & 78.30\% & 82.35\% \\ \hline
Gemini 2.5 Flash & 25.98\% & 41.18\% & 77.94\% & 81.66\% \\ \hline
Claude Sonnet 4 & 23.04\% & 38.24\% & 77.40\% & 81.42\% \\ \hline
\end{tabular}
\end{center}

\textbf{Models can't calculate tax returns reliably today.}

\begin{itemize}
\item Even the best-performing model (Gemini 2.5 Pro) scores only in the mid-30\% range for Correct returns.

\item  Models are inconsistent in their calculations - This is not acceptable for a task which needs consistently correct results with clear auditability. Scores reliably decrease as we increase k in the pass\string^k metric.

\item  While frontier models can calculate some of the simplest returns, they reliably fail to calculate some parts of tax law, e.g. the Child Tax Credit or Earned Income Tax Credit which include complex eligibility requirements.
\end{itemize}

There are some bright spots:

\begin{itemize}
\item  Models do better on the lenient metric: for many returns, the models are only a few dollars off on some lines. This is mostly due to the tax calculation, which in reality relies on a large lookup table (discussed in more detail in the Results Analysis below), but models are often using bracket-based percentage calculations instead, leading to small discrepancies.

\item  Models are better than their overall correct return results on a per line basis: this indicates that there are often single mistakes (discussed in more detail in the Results Analysis below) on the tax return that cascade throughout the rest of the lines, leading to incorrect returns overall.
\end{itemize}

The prompt matters. As part of this experiment, we experimented with prompting to find a prompt we thought to be fair for evaluating models' performance. We landed on a prompt with the following features:

\begin{itemize}
\item  Instructions that the model is helping test tax calculation software: this is because at the time of testing, model safeguards by-default would sometimes refuse to prepare/calculate what it believed to be a real tax return

\item  Instructions to calculate the main Form 1040 and any necessary forms/schedules

\item  Ability to skip the SSN field for ``privacy'' (again, to ensure the model did not refuse for privacy/security safeguards)

\item  A full explanation of the desired output format including line-by-line instructions for the Form 1040

\item  An explanation of the data input format
\end{itemize}

\subsection{Result analysis}

Models fail to correctly compute tax returns in two main ways:

\begin{enumerate}
\item  Using tax bracket percentage-based calculations instead of proper lookup tables (15-20\% of test cases)

\item  Calculation errors
\end{enumerate}

\textbf{Using tax bracket percentage-based calculations}

The 15-20\% delta between strictly Correct returns and Correct returns scored on the lenient metric\break (+/- \$5 difference allowed per-line) is often based on models using percentage, bracket-based calculations for Line 16 of the Form 1040 tax return. This is the line that computes your total tax liability for the year.

Percentage, bracket-based calculations are the commonly-thought-of method for computing tax liability: e.g. M\% of your first \$Nk of income + O\% of your next \$Pk of income. But in reality, \href{https://www.irs.gov/instructions/i1040gi#en_US_2024_publink24811vd0e10588}{the IRS instructions for Form 1040} clearly state ``If your taxable income is less than \$100,000, you must use the Tax Table, later in these instructions, to figure your tax. Be sure you use the correct column. If your taxable income is \$100,000 or more, use the Tax Computation Worksheet right after the Tax Table.''

So models are often ignoring this instruction by-default. For example, in \href{https://github.com/column-tax/tax-calc-bench/blob/main/tax_calc_bench/ty24/results/hoh-multiple-w2-box12-codes/anthropic/claude-opus-4-20250514/model_completed_return_high_1.md?plain=1#L58}{this example run} of the \href{https://github.com/column-tax/tax-calc-bench/tree/main/tax_calc_bench/ty24/test_data/hoh-multiple-w2-box12-codes}{\texttt{hoh-multiple-w2-box12-codes} test case}, Claude Opus 4 calculates Line 16 as:

\begin{codeblock}
Line 15: Subtract line 14 from line 11. If zero or less, enter -0-.
This is your taxable income | $47,900 - $21,900 | 26000
Line 16: Tax | Tax on $26,000 for HOH: $16,550 × 10% + $9,450
× 12% | 2789
\end{codeblock}

which is \href{https://github.com/column-tax/tax-calc-bench/blob/main/tax_calc_bench/ty24/results/hoh-multiple-w2-box12-codes/anthropic/claude-opus-4-20250514/evaluation_result_high_1.md?plain=1#L7}{graded as incorrect by the evaluator} as compared to the proper XML because it's off by \$3:

\begin{codeblock}
Line 16: ✗ incorrect, expected: 2792.0, actual: 2789.0
\end{codeblock}


In reality, the IRS is expecting the use of the Tax look up Table, finding the exact Tax amount for taxpayers with a Head of Household filing status (as in this example) with between \$26,000 and \$26,050 in taxable income:

\begin{center}
  \includegraphics[width=.9\linewidth]{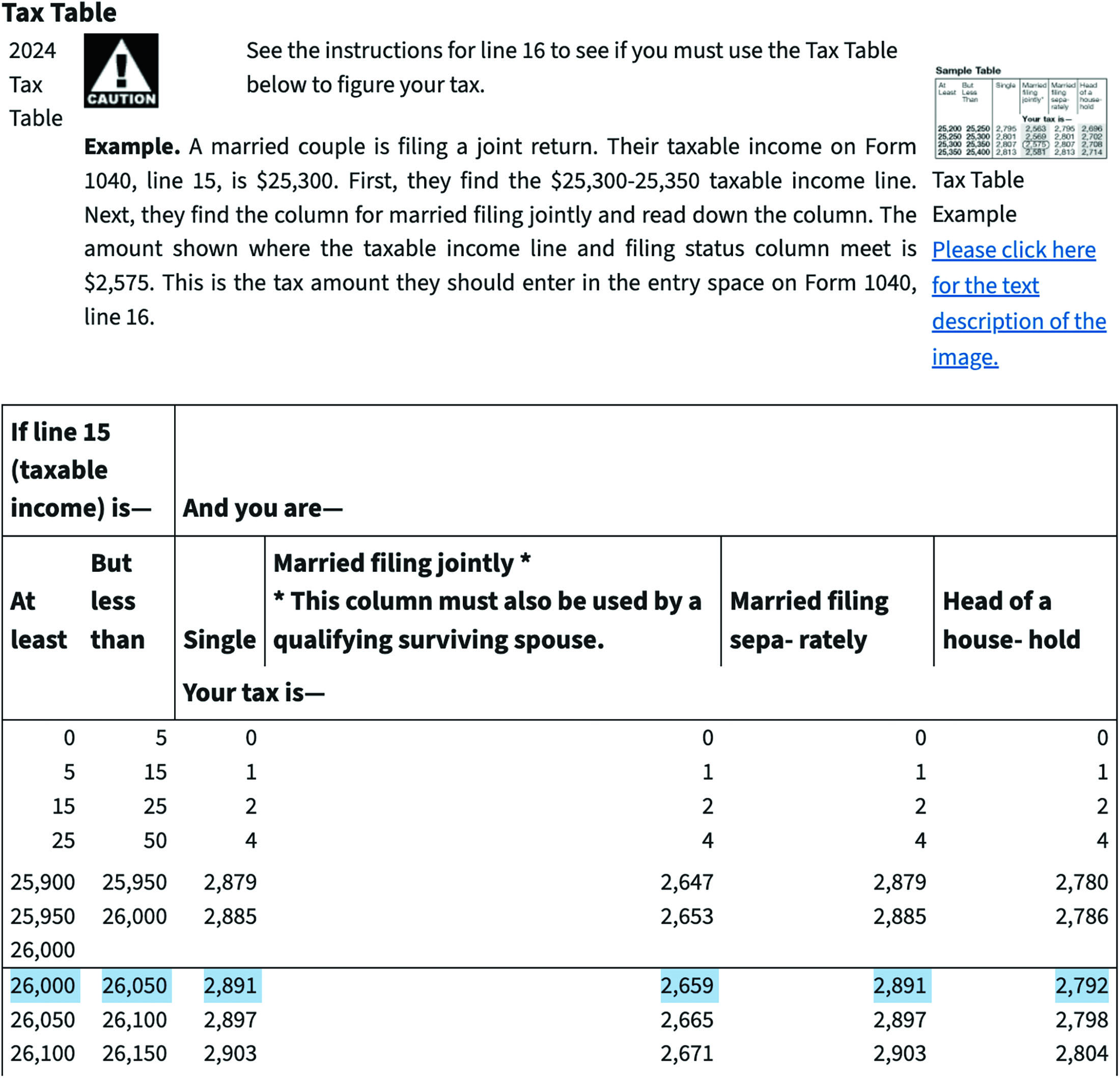}
\end{center}

We can think of methods that could augment the models in order to force them to use the proper Tax Tables to improve on this common error case.

\textbf{Calculation errors}

The majority of return completions attempted by the models have some sort of calculation error. Here is one prototypical example:

On \href{https://github.com/column-tax/tax-calc-bench/blob/main/tax_calc_bench/ty24/results/single-w2-healthcare-marketplace-1095a/gemini/gemini-2.5-flash-preview-05-20/model_completed_return_ultrathink_1.md?plain=1}{this run} of the the \href{https://github.com/column-tax/tax-calc-bench/tree/main/tax_calc_bench/ty24/test_data/single-w2-healthcare-marketplace-1095a}{\texttt{single-w2-healthcare-marketplace-1095a} test case}, Gemini 2.5 Flash, with an ``ultrathink'' (the highest) thinking budget incorrectly computes Form 8962 (for the Premium Tax Credit), leading to an incorrect Schedule 2 and incorrect return starting at Line 17 where the Schedule 2 is pulled back onto the Form 1040.

At this highest thinking budget, Gemini 2.5 Flash \href{https://github.com/column-tax/tax-calc-bench/blob/main/tax_calc_bench/ty24/results/single-w2-healthcare-marketplace-1095a/gemini/gemini-2.5-flash-preview-05-20/model_completed_return_ultrathink_1.md?plain=1}{attempts to break out and calculate} the required Form 8962 and Schedule 2 (Gemini 2.5 Flash often won't do this at lower thinking levels), but then miscomputes the Form.

The model hallucinates incorrect line numbers on Form 8962:

\begin{lstlisting}
Form 8962: Premium Tax Credit (PTC)
===================================
Line 1: Annual household income | Your adjusted gross income | 28,125
Line 2: Household size | Total number of individuals in your tax
household | 1
Line 3: Household income as a percentage of federal poverty line (FPL)
 | Line 1 / FPL for your household size ($15,060) | 186.75%
\end{lstlisting}

Lines 1 and 2 are incorrectly swapped, and Line 3 should be ``Household income. Add the amounts on lines 2a and 2b. See instructions'' not ``Household income as a percentage of federal poverty line''.\eject

Here is the correct Form for comparison:

\begin{figure}[h!]
  \centering
  {\includegraphics[width=.9\linewidth]{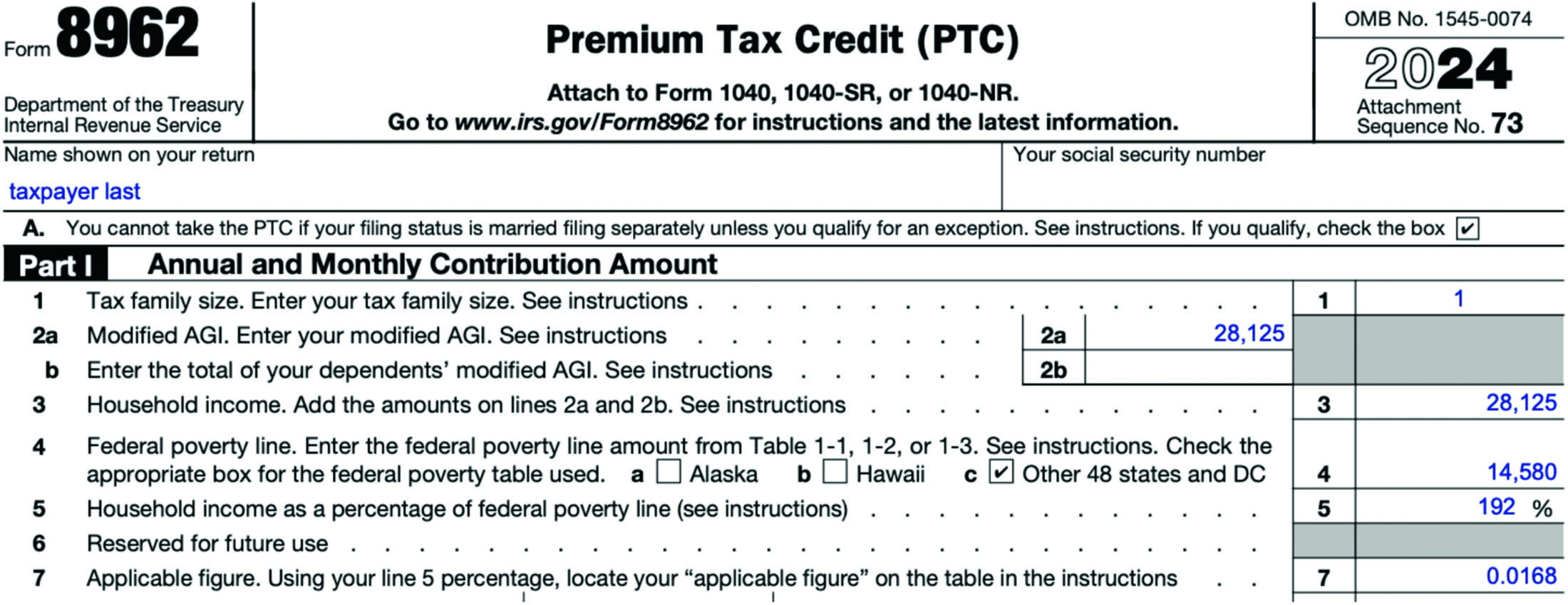}}
\end{figure}

The model also uses the wrong Federal Poverty Level (FPL): \$15,060 vs \$14,580 (which caused additional mistakes down the form).

Mistakes of this nature continue through the run. For example, Gemini 2.5 Flash computed Schedule 2 as:

\begin{lstlisting}
Form Schedule 2: Additional Taxes
=================================
Line 1: Tax from Form 1040, line 16 | Reference to Form 1040, line 16
 | 1,391
Line 2: Excess advance premium tax credit repayment | From Form 8962,
line 27 | 158
Line 3: Add lines 1 and 2 | Sum of lines 1 and 2 ($1,391 + $158) |
1,549
Line 21: Add lines 11 through 20. These are your total additional
taxes | No other additional taxes apply | 0
\end{lstlisting}

Which doesn't match the official Schedule 2 at all:

\begin{figure}[hb!]
  \centering
  \vspace*{12pt}
  {\includegraphics[width=.9\linewidth]{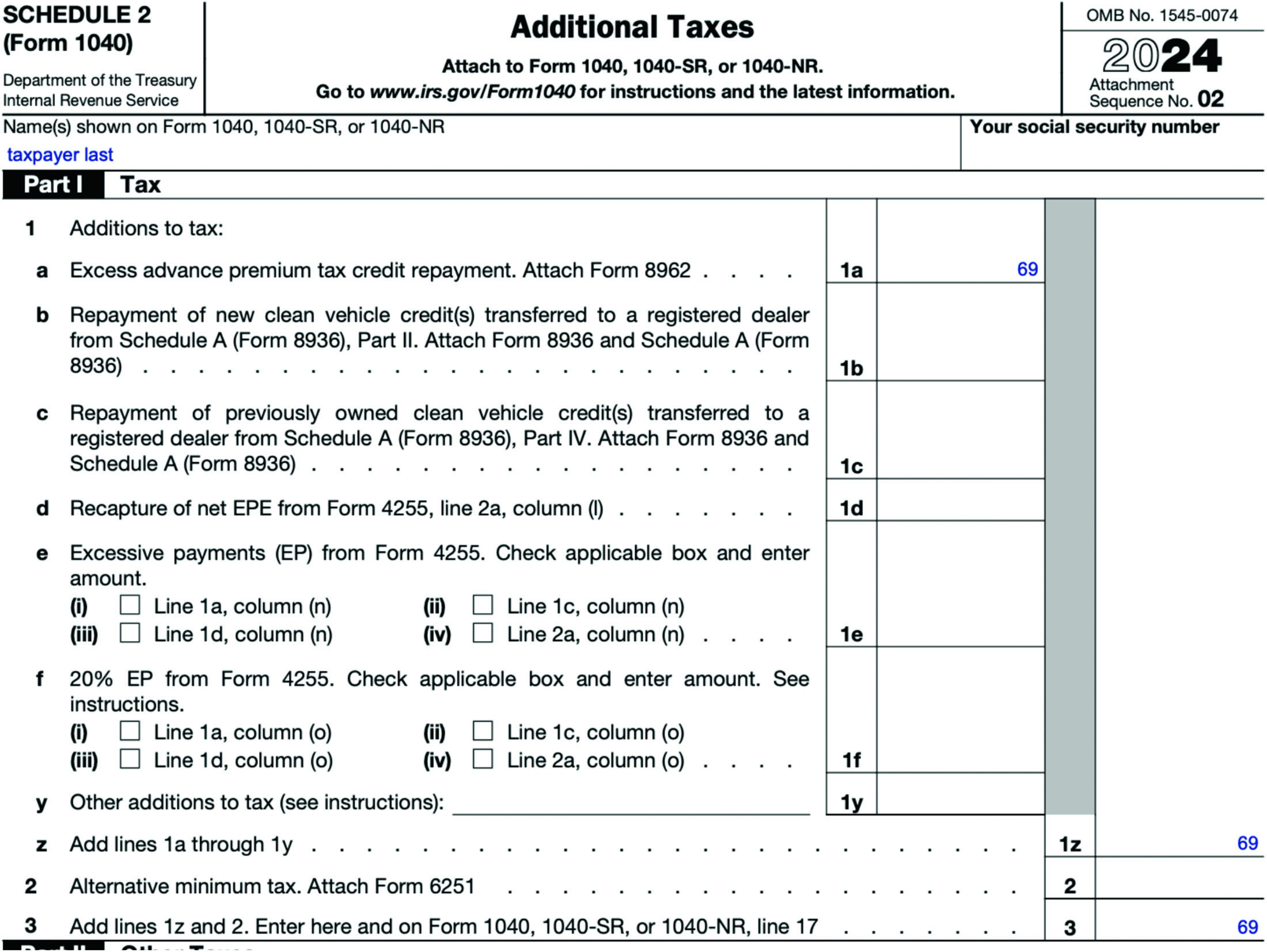}}
\end{figure}
\vfill\eject

These results point to the need for additional scaffolding, infrastructure, and augmentation that could potentially help the models compute some of the additional forms. For example, in the \href{https://github.com/column-tax/tax-calc-bench/blob/main/tax_calc_bench/tax_return_generation_prompt.py}{prompt} for this experiment, the model is bootstrapped with the correct Form 1040 lines and short instructions as part of its context. We imagine similar techniques could help with additional forms, but getting this right will be tricky.

\subsection{Per-provider takeaways}

We only tested models that have a 2025 knowledge cutoff because those are models which have complete information about the 2024 tax year. If you're a model provider looking to test your model on this benchmark, feel free to \href{mailto:team@columntax.com}{contact us} for help.

\textbf{Gemini}

Gemini 2.5 Pro is the best-performing model on this benchmark.

\begin{itemize}
\item  Interestingly, model performance does not increase for Gemini 2.5 Pro above a certain thinking budget. This indicates that above that thinking budget, the model is not spending its thinking tokens on anything that can improve its performance.

\item  By default, Gemini's API includes a \href{https://ai.google.dev/gemini-api/docs/thinking#set-budget}{dynamic thinking} budget for its 2.5 Pro and 2.5 Flash models. This works well for the tax calculation task, which for the 2.5 Flash model, requires at least some thinking budget to get improved performance.
\end{itemize}

\begin{figure}[h!]
  \centering
  {\includegraphics[width=.55\linewidth]{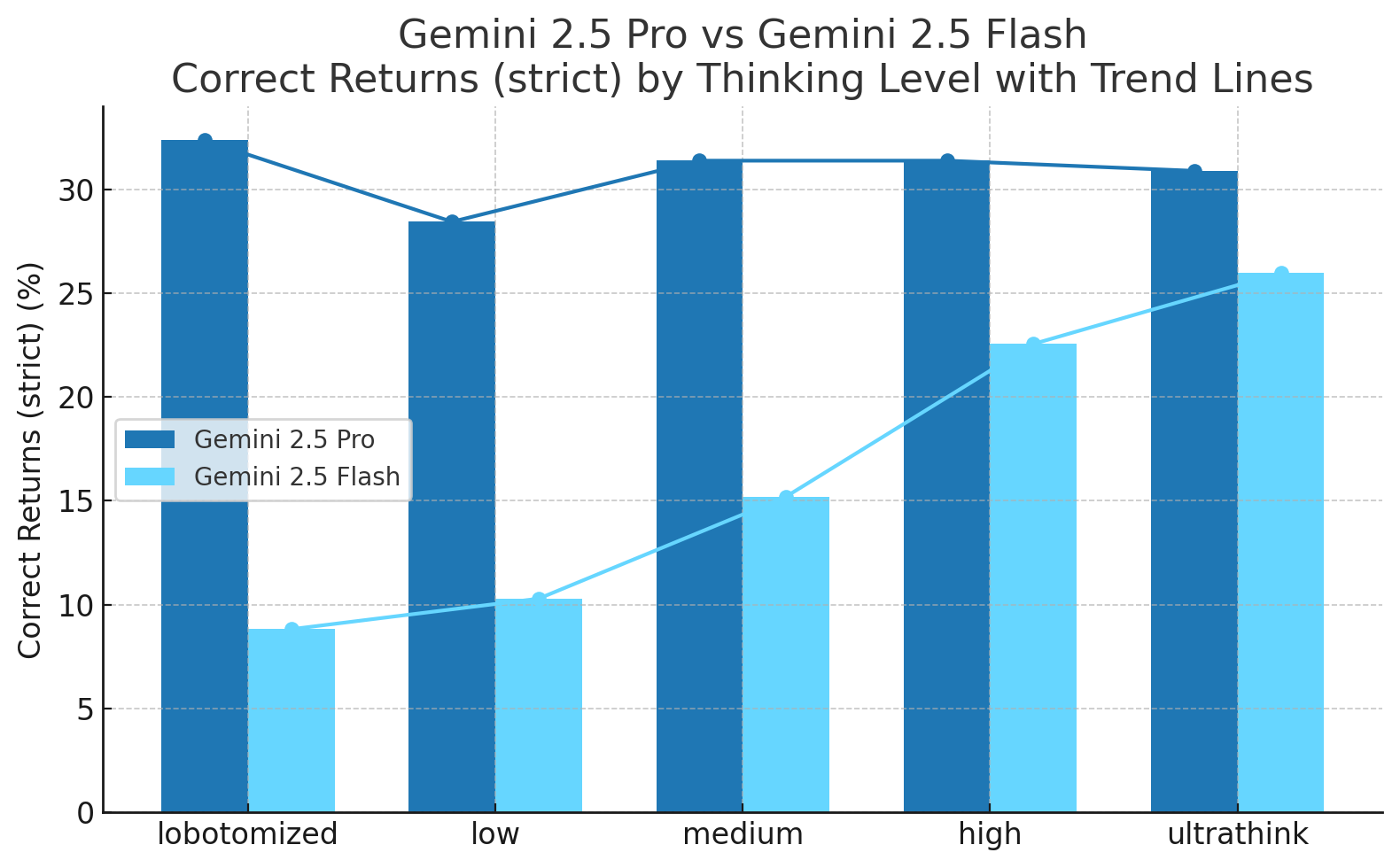}}
\end{figure}

\textbf{Claude}

Claude Opus 4 is the second best-performing model in this benchmark, but still lags in performance behind Gemini 2.5 Pro at lower thinking budget levels.

\begin{enumerate}
\item  Claude's Opus and Sonnet models see greatly improved performance with increased thinking budgets.

\item  By default, \href{https://docs.anthropic.com/en/docs/build-with-claude/extended-thinking}{Claude's API has thinking budgets disabled}, which significantly hampers Claude's performance on this benchmark.
\end{enumerate}

\begin{figure}[h!]
  \centering
  {\includegraphics[width=.55\linewidth]{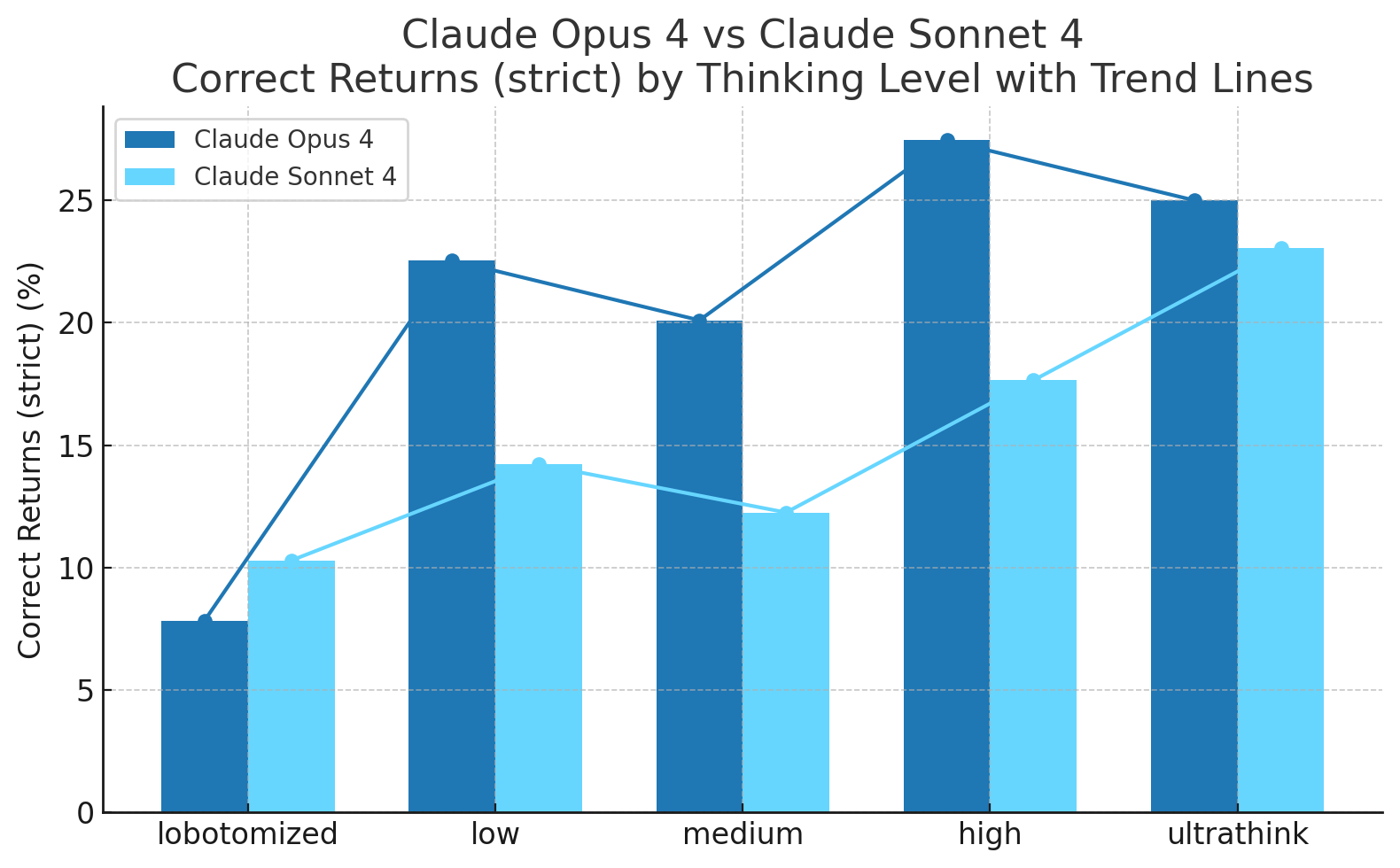}}
\end{figure}

\subsection{Detailed results}

{\renewcommand{\arraystretch}{1.5}
\begin{tabular}{|p{2.5cm}|c|c|c|c|c|c|}
\hline
\textbf{Model Name} & \textbf{Thinking} & \textbf{Tests Run} & \parbox{1.3cm}{\centering\textbf{Correct Returns (strict)}} & \parbox{1.3cm}{\centering\textbf{Correct Returns (lenient)}} & \parbox{1.3cm}{\centering\textbf{Correct (by line)}} & \parbox{1.3cm}{\centering\textbf{Correct (by line, lenient)}} \\
\hline
gemini-2.5-pro-preview-05-06 & lobotomized & $51{\times}4/51$ & 32.35 \% & 51.96 \% & 80.91 \% & 85.86 \% \\ \hline
gemini-2.5-pro-preview-05-06 & high & $51{\times}4/51$ & 31.37 \% & 51.47 \% & 81.22 \% & 86.12 \% \\ \hline
gemini-2.5-pro-preview-05-06 & medium & $51{\times}4/51$ & 31.37 \% & 51.47 \% & 80.26 \% & 85.17 \% \\ \hline
gemini-2.5-pro-preview-05-06 & ultrathink & $51{\times}4/51$ & 30.88 \% & 50.49 \% & 80.03 \% & 84.93 \% \\ \hline
gemini-2.5-pro-preview-05-06 & low & $51{\times}4/51$ & 28.43 \% & 49.02 \% & 79.95 \% & 84.75 \% \\ \hline
claude-opus-4-20250514 & high & $51{\times}4/51$ & 27.45 \% & 42.65 \% & 78.30 \% & 82.35 \% \\ \hline
gemini-2.5-flash-preview-05-20 & ultrathink & $51{\times}4/51$ & 25.98 \% & 41.18 \% & 77.94 \% & 81.66 \% \\ \hline
claude-opus-4-20250514 & ultrathink & $51{\times}4/51$ & 25.00 \% & 41.18 \% & 77.43 \% & 81.94 \% \\ \hline
claude-sonnet-4-20250514 & ultrathink & $51{\times}4/51$ & 23.04 \% & 38.24 \% & 77.40 \% & 81.42 \% \\ \hline
claude-opus-4-20250514 & low & $51{\times}4/51$ & 22.55 \% & 37.75 \% & 77.37 \% & 81.32 \% \\ \hline
gemini-2.5-flash-preview-05-20 & high & $51{\times}4/51$ & 22.55 \% & 36.76 \% & 75.21 \% & 79.31 \% \\ \hline
claude-opus-4-20250514 & medium & $51{\times}4/51$ & 20.10 \% & 35.78 \% & 76.08 \% & 80.11 \% \\ \hline
claude-sonnet-4-20250514 & high & $51{\times}4/51$ & 17.65 \% & 25.00 \% & 74.79 \% & 77.24 \% \\ \hline
gemini-2.5-flash-preview-05-20 & medium & $51{\times}4/51$ & 15.20 \% & 25.49 \% & 70.49 \% & 73.63 \% \\ \hline
claude-sonnet-4-20250514 & low & $51{\times}4/51$ & 14.22 \% & 21.57 \% & 73.63 \% & 76.24 \% \\ \hline
claude-sonnet-4-20250514 & medium & $51{\times}4/51$ & 12.25 \% & 20.59 \% & 73.22 \% & 75.95 \% \\ \hline
gemini-2.5-flash-preview-05-20 & low & $51{\times}4/51$ & 10.29 \% & 19.12 \% & 69.30 \% & 72.70 \% \\ \hline
claude-sonnet-4-20250514 & lobotomized & $51{\times}4/51$ & 10.29 \% & 12.25 \% & 70.07 \% & 71.57 \% \\ \hline
gemini-2.5-flash-preview-05-20 & lobotomized & $51{\times}4/51$ & 8.82 \% & 11.27 \% & 66.80 \% & 68.27 \% \\ \hline
claude-opus-4-20250514 & lobotomized & $51{\times}4/51$ & 7.84 \% & 11.27 \% & 70.61 \% & 72.47 \% \\ \hline
\end{tabular}}

The Tests Run column shows tests${\times}$runs/total (e.g., $51{\times}4/51$ means 51 test case runs 4 times each of 51 total test cases).\vfill\eject

\begin{figure}[h!]
  \centering
  {\includegraphics[width=.95\linewidth]{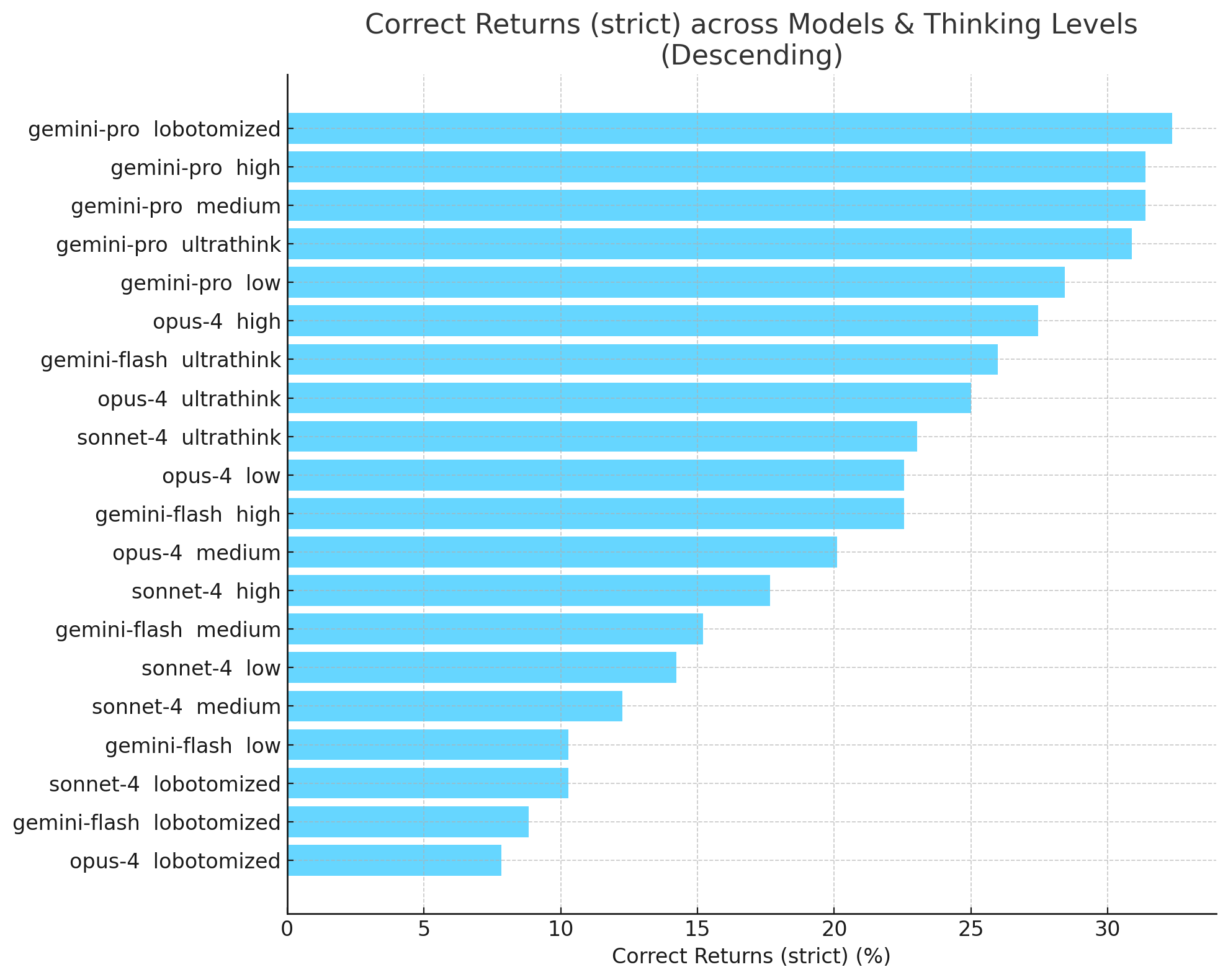}}
\end{figure}

Models are not consistent in their calculations today, as seen via the pass\string^k metric decreasing as k increases:

\begin{figure}[h!]
  \centering
  {\includegraphics[width=.95\linewidth]{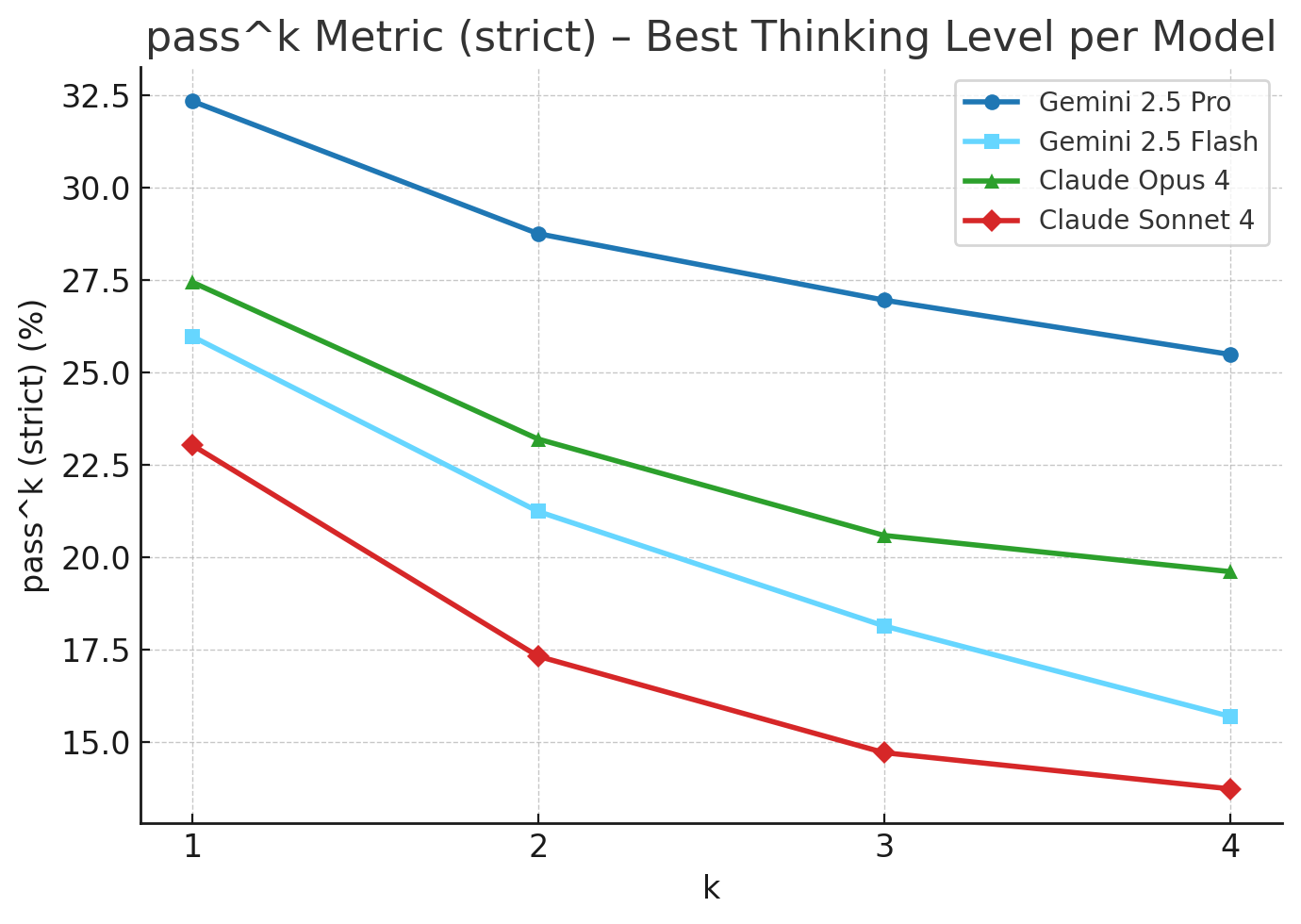}}
\end{figure}

\section{The goal posts will move}

The TY24 edition of TaxCalcBench is a slimmed-down version of the true complexity of the task:

\begin{itemize}
\item  The dataset is federal-only (42 states + D.C. levy state/local income tax)

\item  It covers only a relatively simple set of tax situations: the vast majority of tax forms are not covered by this dataset

\item  It does not expect the output to be formatted in \href{https://www.irs.gov/e-file-providers/modernized-e-file-mef-schemas-and-business-rules}{MeF schema}-compatible XML
\end{itemize}

We expect to release a yearly version of the benchmark and for future editions to add state returns, more-complex situations, and to switch to testing against proper XML output.

\section{Discussion}

\textbf{Further work}

\begin{itemize}
\item  We'd like to spend additional time analyzing why additional thinking budget does not increase performance for the top-tier Gemini 2.5 Pro model. Our initial hypothesis is that the model either confuses itself further or has no resources to spend its thinking budget on anything additionally useful.

\item  We'd like to spend additional time on prompt engineering to understand if there's more we can do to reduce failures from the model on both the tax table usage and calculation errors.

\item  We'd like to experiment with the temperature setting of models and that setting's impact on the pass\string^k metric in particular.

\item  As discussed in the section above, in future interactions of the benchmark, we'd like to extend the dataset to include more-complex returns (e.g. covering the longer tail of income situations and tax law) as well as extend to include state tax returns.

\item  Future iterations of the benchmark will require models to output the IRS-required XML format rather than a simplified markdown format.
\end{itemize}

\section{Related Work}

Since there are so few available tax engines in the US and all of them belong to commercial entities, there is very little publicly-available research on the topic. To our knowledge, this is the first AI benchmark of this type for the tax calculation task.

This work is inspired by other benchmarks, in particular HumanEval \cite{6} for its use of the pass@k metric and TAU-bench \cite{7} for the introduction of the pass\string^k metric.

\section*{Acknowledgments}
This benchmark dataset would not have been possible without the hard work of the Tax Analyst team at Column Tax who has worked tirelessly over the past 4 years to create one of the first at-scale tax calculation engines in the past two decades and whose testing work is the underpinning and source of the data in this benchmark.

\bibliography{bib}
\bibliographystyle{unsrt}

\appendix

\section{Additional analysis}

Another run that serves as a good example of models' typical performance on this task is \href{https://github.com/column-tax/tax-calc-bench/blob/main/tax_calc_bench/ty24/results/single-senior-blind-over-65/anthropic/claude-sonnet-4-20250514/model_completed_return_high_4.md}{the 4th run} of Claude Sonnet 4 at high thinking level on the \texttt{single-senior-blind-over-65} test case.

In this test case, Claude Sonnet 4 correctly identified the taxable interest amount to be reported, but incorrectly reduced the taxable interest by the penalty on early withdrawal of savings that should instead be reported on Schedule 1, line 18:

\begin{lstlisting}
Line 2b: Taxable interest | Interest income less early withdrawal
penalty: $1,222 - $1,212 | 10
\end{lstlisting}

Compared to the correctly-outputted Form 1040 and Schedule 1:

\begin{figure}[H]
  \centering\vspace*{12pt}
  {\includegraphics[width=\linewidth]{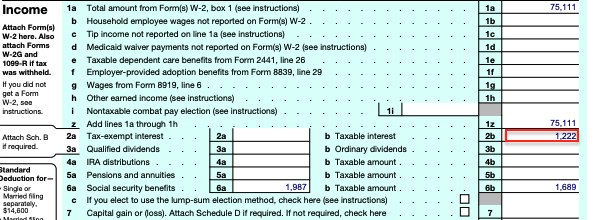}}
\end{figure}

\begin{figure}[H]
  \centering
  {\includegraphics[width=\linewidth]{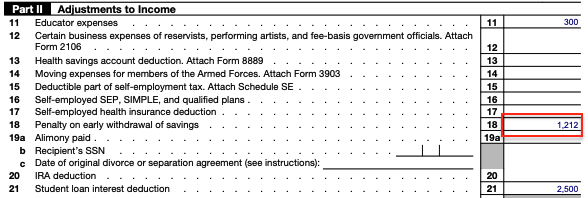}}
  \vspace*{0pt}
\end{figure}

On the other hand, Claude Sonnet 4 performs well at calculating the taxable social security benefits:

\begin{lstlisting}[language=Python]
Line 6b: Taxable amount | 85% of benefits due to income level | 1,690
\end{lstlisting}

\vfill\eject

Which is only off by \$1 from the expected amount, possibly due to a rounding issue:

\begin{figure}[H]
  \centering
  {\includegraphics[width=\linewidth]{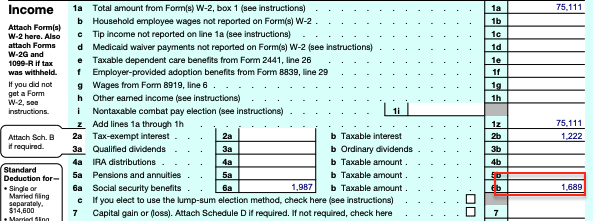}}
\end{figure}

While Claude Sonnet 4 here outputs an explanation of ``85\% of benefits due to income level'' - the full calculation relies on the worksheet in \href{https://www.irs.gov/pub/irs-pdf/p915.pdf}{IRS Publication 915} which has a more-involved multi-party calculation.
\end{document}